\documentclass[conference]{IEEEtran}
\usepackage{times}
\usepackage[T1]{fontenc}
\usepackage[utf8]{inputenc}
\usepackage[numbers]{natbib}
\usepackage{amsmath}
\usepackage{amssymb}
\usepackage{booktabs}
\usepackage{graphicx}
\usepackage{wrapfig}
\usepackage{xcolor}
\usepackage{pifont}
\usepackage{multirow}
\usepackage{tabularx}
\usepackage{array}
\usepackage{colortbl}
\usepackage{float}
\usepackage{placeins}
\usepackage{url}
\usepackage[bookmarks=true]{hyperref}

\definecolor{highlight}{RGB}{30,90,160}
\definecolor{percentcolor}{RGB}{30,90,160}
\newcommand{\cmark}{\ding{51}}
\newcommand{\xmark}{\ding{55}}
\begin{document}
\title{Reducing Temporal Redundancy for Efficient Vision-Language-Action Inference}

\author{
\IEEEauthorblockN{
\parbox{\textwidth}{
\centering
Yuzhou Wu$^{1,*}$,
Yuxin Zheng$^{2,*}$,
Muchun Niu$^{3,*}$,
Yishan Yang$^{4}$,
Tianhao Liu$^{1}$
\\[3pt]
Hanwen Kang$^{1}$,
Jiajian Jing$^{6}$,
Linfeng Zhang$^{3,\dagger}$,
Chuan Wen$^{3,\dagger}$
}
}

\IEEEauthorblockA{
\parbox{\textwidth}{
\centering
$^{1}$Tianji KernalMind co ltd\\
$^{2}$Tsinghua Shenzhen International Graduate School,\\
Tsinghua University, Shenzhen, China\\
$^{3}$Shanghai Jiao Tong University, Shanghai, China\\
$^{4}$Southeast University, Nanjing, China\\
$^{6}$Huazhong Agricultural University, Wuhan, China\\[3pt]
$^{*}$Equal contribution
\qquad
$^{\dagger}$Corresponding authors
}
}
}

\maketitle

\begin{abstract}
    Vision-Language-Action (VLA) models exhibit strong generalization for robotic manipulation, yet their high inference latency limits real time deployment. We identify two primary sources of temporal redundancy in existing VLA pipelines: repeated visual encoding of highly similar consecutive frames and multi step iterative sampling in diffusion based policies. To address this, we propose a system level acceleration strategy that reduces computation in both perception and action generation. On the perception side, we incrementally update only tokens corresponding to dynamic scene regions instead of re-encoding entire frames. On the policy side, we compress diffusion sampling into a compact 2-step schedule through efficiency oriented training while preserving action precision. Experiments on Libero, RobotWin, and Real Robot Platforms demonstrate over $2\times$ speedup while maintaining high performance, achieving up to 98\% success rate on general manipulation benchmarks. Our codes will be released on Github.
\end{abstract}

\IEEEpeerreviewmaketitle
\begin{figure*}[t]
\centering
\includegraphics[width=\textwidth]{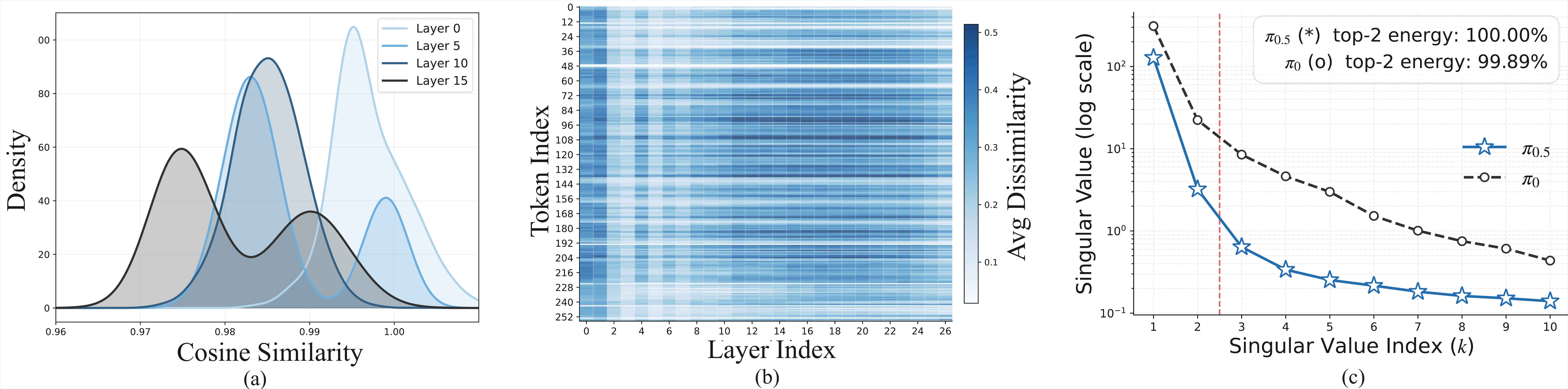}
\caption{
\textbf{Structural redundancy analysis in perception and policy dynamics.}
(a) Distribution of cosine similarity between visual tokens of consecutive frames across ViT layers. Most tokens exhibit near unity similarity ($>0.98$), revealing strong temporal redundancy in streaming robotic perception. (b) Evolution of token dissimilarity across layers. Only a sparse subset of spatial tokens show significant deviation, supporting selective token refresh. (c) Intrinsic dimension analysis of flow matching velocity fields. Singular value spectra indicate that most energy is concentrated in the top two components, suggesting that velocity evolution lies in a low dimensional subspace. The rapid decay of lower index singular values implies strong linear correlation among velocity updates across time steps, allowing multiple diffusion steps to be approximated with a small number of updates. Together, these observations motivate our approach to exploit temporal redundancy in both perception and policy, enabling efficient VLA inference through selective token reuse and step-compressed policy execution.
}
\label{fig:overview}
\end{figure*}
\section{Introduction}

The rapid advancement of Vision-Language-Models (VLMs) has driven the emergence of general purpose Vision-Language-Action (VLA) agents capable of interpreting natural language instructions and executing multi step physical tasks \cite{ma2024survey, awadalla2023openflamingo, li2022blip, radford2021learning}. Systems such as RT-2, OpenVLA, and $\pi_0$, built on Internet scale multimodal pre training, have significantly advanced robotic autonomy \cite{zitkovich2023rt, kim2024openvla, black2024pi}. However, a substantial gap still remains between their impressive capabilities and real world deployment. Modern VLA pipelines rely heavily on large Vision Transformer (ViT) backbones and iterative inference mechanisms such as diffusion or flow matching \cite{han2022survey, croitoru2023diffusion, lipman2022flow}, introducing considerable latency. This latency directly conflicts with the high frequency, closed loop control required for safe and responsive manipulation in dynamic environments \cite{zhao2023learning, fu2024mobile}. As a result, efficient inference is no longer merely desirable but absolutely essential for practical robotic deployment \cite{zitkovich2023rt}.

The need for low latency inference has motivated a growing body of research on VLA acceleration. However, most existing approaches focus on optimizing isolated components, resulting in limited system level gains. This fragmented strategy often shifts the computational bottleneck rather than eliminating it. For instance, methods such as TinyVLA \cite{wen2025tinyvla} and DeeR-VLA \cite{yue2024deer} design lightweight architectures but are difficult to apply to large pre-trained models. Meanwhile, approaches like VLA-Cache \cite{xu2025vla} exploit visual redundancy through token caching, yet primarily target the perception module. The memory overhead of the LLM core and the iterative computation of the action head remain largely unaddressed. Furthermore, many existing acceleration studies are evaluated mainly in simulated environments and concentrate on improving VLM components, leaving the efficiency of action generation largely unexplored. Consequently, current approaches fall short of addressing the end to end latency of modern VLA agents at the system level\cite{tang2025vlash, tan2025think, pertsch2025fast}.
\begin{figure}[t]
\centering
\includegraphics[width=\linewidth]{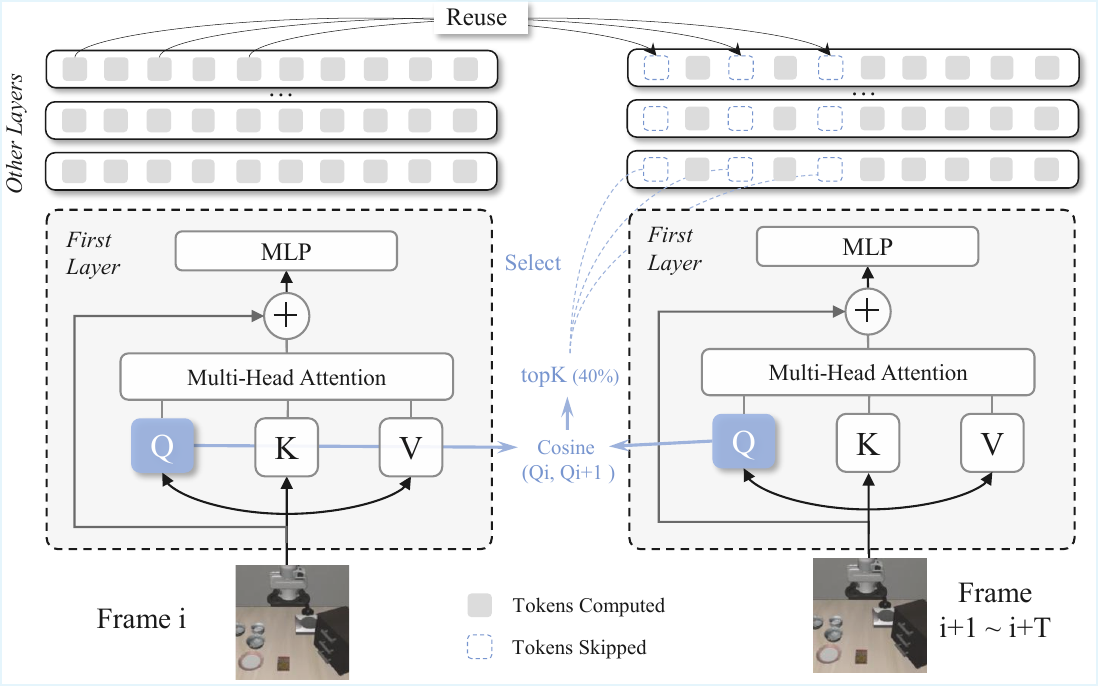}
\caption{
\textbf{Temporal token reuse with cosine similarity selection.}
Only a subset of tokens is updated based on cosine similarity, while others reuse cached representations.
Index selection is performed in the first layer and reused across subsequent layers.
}
\label{fig:token_reuse}
\end{figure}
To understand the source of inefficiency in VLA inference, we analyze the temporal behavior of both perception and policy modules. As shown in Fig.~\ref{fig:overview}, the perception side Fig.~\ref{fig:overview}(a,b) reveals a strong temporal consistency in visual representations. Visual tokens extracted from adjacent frames exhibit extremely high cosine similarity across multiple ViT layers, with most values consistently above 0.98. In addition, token dissimilarity maps show clear spatial sparsity: only a small fraction of tokens change noticeably between steps, while the majority remain nearly unchanged over time. These results indicate substantial temporal redundancy in streaming robotic perception. On the policy side Fig.~\ref{fig:overview}(c), we examine the singular value spectrum of velocity trajectories in flow matching. The spectrum decays rapidly, with energy concentrated in the top two dominant components. This suggests that velocity evolution follows a low dimensional structure. As a result, multiple diffusion iterations mainly produce highly correlated refinements rather than independent directional updates.

Taken together, these findings expose a systemic structural redundancy across the entire VLA pipeline, motivating a unified acceleration strategy rather than isolated module level optimizations. Building directly upon the above observations, we propose a unified system level acceleration framework for flow based VLA models that reduces temporal redundancy in both perception and action generation. On the perception side, instead of re-encoding entire frames at every step, we incrementally update only tokens corresponding to dynamic regions between adjacent frames, effectively preserving static representations through token reuse; empirically, most scenes require recomputing only about 60\% of tokens per frame. On the policy side, we exploit the low rank, linearly correlated structure of flow velocities and reconstruct the diffusion inference process into a compact two step schedule through efficiency oriented training, significantly reducing iterative computation while maintaining trajectory precision in practice.

Extensive experiments on LIBERO, RobotWin, and Real World Robotic Platforms demonstrate that our approach achieves more than $2\times$ end to end inference speedup with negligible performance degradation. 
In particular, our method maintains comparable task accuracy while achieving nearly $2.6\times$ end to end speedup in simulation and $1.6\times$ acceleration on Real Robot Platforms.  These results demonstrate that real time deployment of large scale VLA models is attainable without compromising manipulation performance. Our main contributions are summarized as follows:
\begin{itemize}
    \item We identify previously overlooked structural redundancies in VLA inference, including strong temporal consistency in visual tokens across adjacent frames and low rank velocity dynamics in flow based action generation.
    
    \item Based on these insights, we propose a unified system level acceleration framework for flow based VLA models that reduces perception redundancy across adjacent frames through selective token updates, and reduces policy computation through a compact two-step flow inference scheme.
    
    \item We validate the proposed approach across simulation benchmarks (LIBERO, RobotWin) and real word robotic platforms, achieving over $2\times$ end to end inference speedup while maintaining comparable task performance.
\end{itemize}
\section{Related Work}
\begin{figure}[t]
\centering
\includegraphics[width=\linewidth]{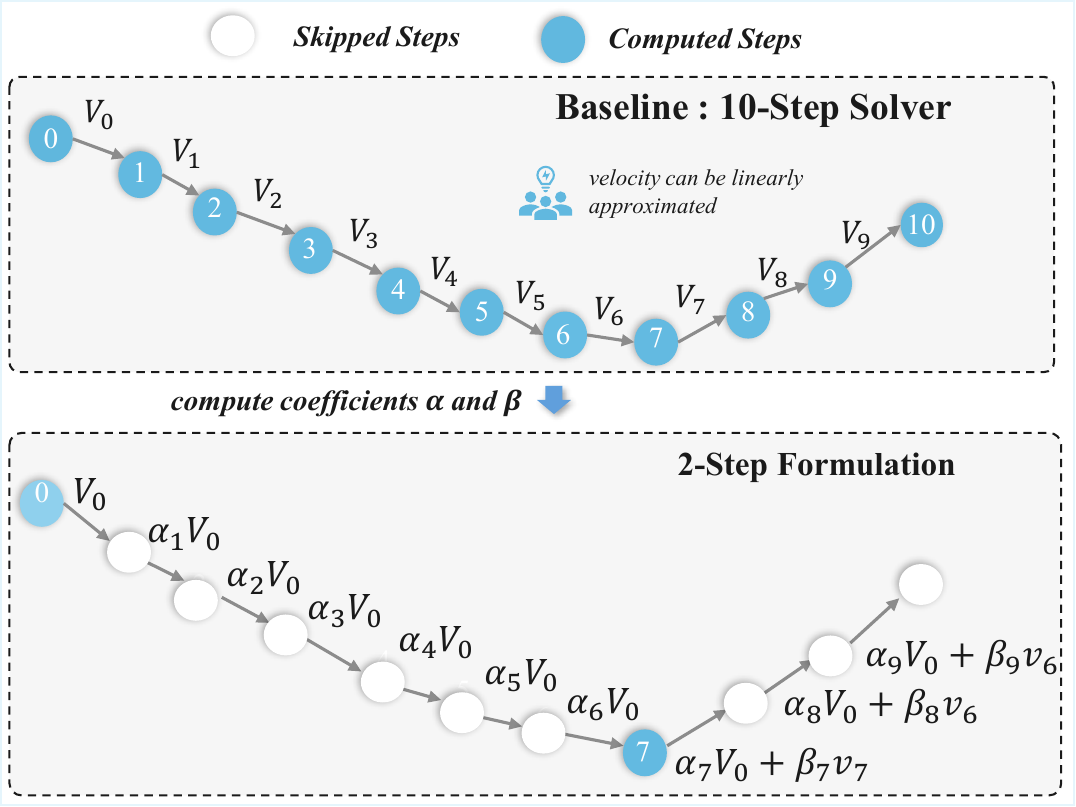}
\caption{
\textbf{Flow step compression via low rank velocity structure.}
}
\label{fig:flow}
\end{figure}
\subsection{Vision-Language-Action Models for Manipulation}

Vision-Language-Action (VLA) models aim to unify perception, language understanding, and control within a single framework, thereby enabling robots to execute diverse tasks specified by natural language instructions~\cite{sapkota2025vision}. Recent works increasingly leverage large scale vision language pre training to endow robots with strong semantic grounding and generalization. Representative systems such as RT-2~\cite{zitkovich2023rt}, OpenVLA~\cite{kim2024openvla}, and $\pi_0$~\cite{black2024pi_0} adopt transformer based architectures and Internet scale data to directly map multimodal observations to actions, demonstrating consistently impressive zero shot and multi task manipulation capabilities.

Despite their success, most existing VLA systems are primarily designed with a focus on task generality and semantic alignment rather than real time efficiency. In practice, these models often rely on heavy vision backbones (e.g., SigLIP or ViT Huge)~\cite{zhai2023sigmoid} and complex policy heads, resulting in high inference latency that hinders effective closed loop control~\cite{wen2025tinyvla}. Our work complements this line of research by directly targeting efficient inference for VLA models without sacrificing their general purpose nature.

\subsection{Efficient Visual Perception for Robotics}

Efficient perception has long been a central topic in robotics, with early works focusing on lightweight convolutional architectures and feature reuse\cite{howard2017mobilenets,zhang2018shufflenet}. More recently, Vision Transformers (ViTs) have become dominant due to their strong representation capacity, but their quadratic complexity with respect to token count makes them computationally expensive for high frequency control\cite{dosovitskiy2020image,liu2021swin}.

Several approaches have explored reducing the cost of ViT inference via token pruning, token merging, or early exiting. While effective in static or single image settings, these methods typically process each frame independently and fail to fully exploit the strong temporal continuity present in robotic video streams. In contrast, our approach explicitly leverages temporal redundancy by reusing visual tokens across consecutive frames and incrementally updating only regions directly affected by scene changes, enabling efficient streaming perception for real time manipulation.

\subsection{Temporal Modeling and Incremental Inference}

Temporal modeling is commonly addressed by recurrent networks, temporal convolutions, or video transformers that jointly process multiple frames\cite{yu2019review}. However, these methods often significantly increase computational cost and memory usage, making them generally less suitable for high frequency online control in practice.

Incremental inference has been explored in areas as video understanding and streaming transformers, where cached representations are reused across time steps. These techniques demonstrate that significant redundancy exists in consecutive inputs. Our method builds on this insight but is tailored to the robotics setting: rather than caching entire frames or sequences, we selectively reuse spatial tokens while maintaining feature consistency through targeted training, ensuring stable control under partial feature updates.

\subsection{Diffusion and Flow Based Policies for Control}

Diffusion and flow matching models have recently emerged as powerful policy representations for continuous control and robotic manipulation, due to their ability to model complex, inherently multi modal action distributions\cite{janner2022planning,lipman2022flow,chi2025diffusion}. Prior works consistently show that iterative denoising processes can produce high quality trajectories, but often at the cost of multiple forward passes per control step, inevitably leading to high latency in practice overall.

To address this limitation, several studies investigate accelerated sampling strategies, such as reducing the number of diffusion steps or distilling iterative processes into fewer steps\cite{pearce2023imitating}. Our work follows this direction but focuses on preserving control precision under aggressive step compression. By combining a compact sampling schedule with an efficiency oriented training recipe, we achieve fast action generation suitable for real time closed loop control\cite{lu2025dpm,salimans2022progressive}.
\section{Methodology}
\begin{figure*}[t]
\centering
\includegraphics[width=\textwidth]{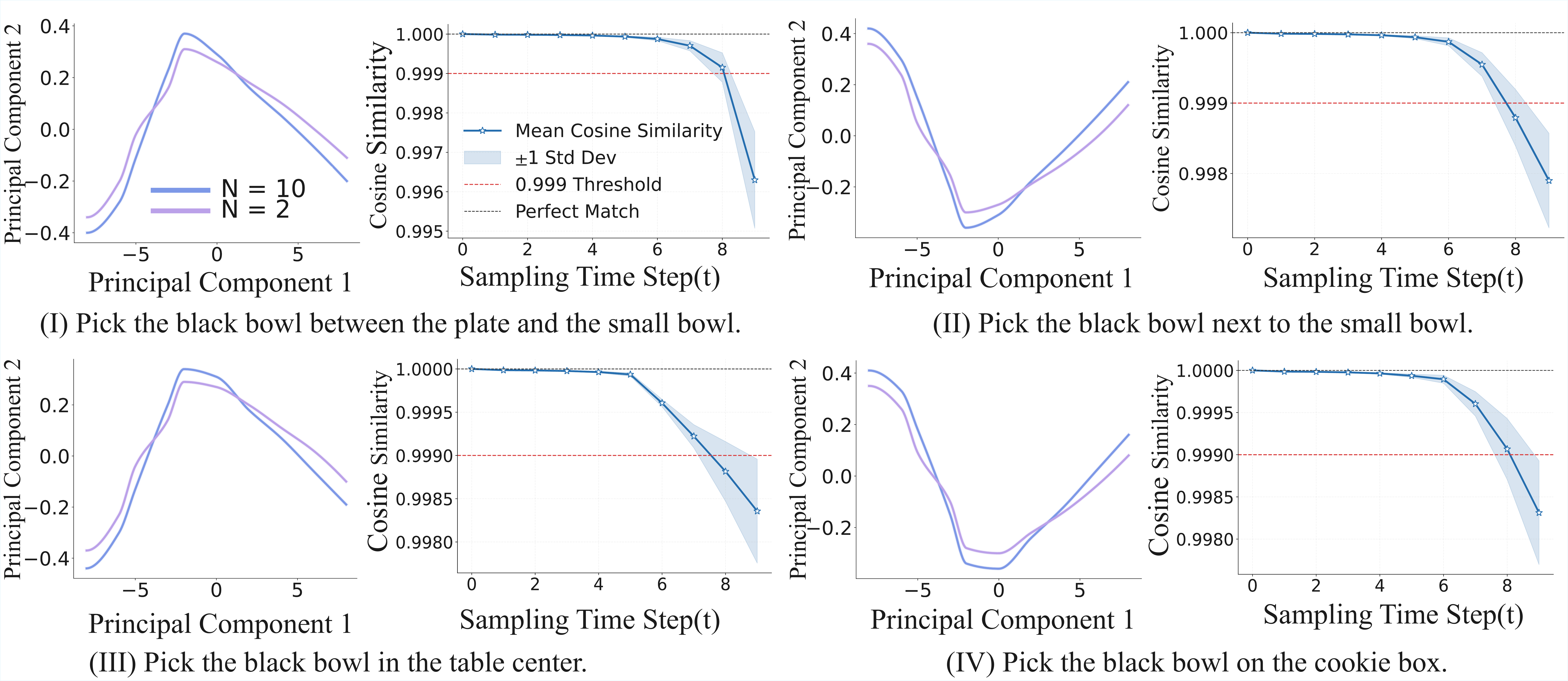}
\caption{
\textbf{Flow matching step compression analysis.}
For four representative manipulation tasks from the Libero Spatial benchmark, we visualize the PCA projected action trajectories together with the cosine similarity between intermediate velocity fields and the initial velocity $v_0$. The compressed 2-step solver produces trajectories that remain geometrically consistent with those generated by the original 10-step solver in the latent space.
Meanwhile, the cosine similarity stays close to one across most sampling steps, revealing strong directional consistency in the velocity dynamics and explaining why aggressive step compression is possible.
}
\label{fig:flow_compression}
\end{figure*}
We propose an efficient acceleration method for Vision-Language-Action (VLA) systems that reduces redundant computation in both perception and action generation. Our method introduces temporally aware visual token reuse and a compressed 2-step flow matching policy, enabling significant latency reduction without altering the backbone architecture. The resulting framework supports real time closed loop control while preserving task performance.

\subsection{Preliminaries: Vision-Language-Action Models}

Vision-Language-Action (VLA) models aim to directly map multimodal observations to continuous control actions. Given an observation $o = (I, L, s)$ consisting of image $I$, language tokens $L$, and robot state $s$, a VLA policy predicts an action sequence $a \in \mathbb{R}^{H \times D}$ over a horizon $H$. Modern VLA systems typically adopt a transformer based architecture composed of: a vision encoder (e.g., SigLIP/ViT) that converts images into spatial tokens, a large language backbone that fuses visual and textual embeddings a policy head that generates actions via diffusion or flow matching inference.

In flow matching based policies, action generation is formulated as solving an ordinary differential equation (ODE):\begin{equation}
\frac{d x_t}{dt} = v_\theta(x_t, t, o),
\end{equation} where $x_t$ denotes the noisy action at time $t \in [0,1]$, and $v_\theta$ is a learned velocity field conditioned on observation.
Inference typically starts from Gaussian noise $x_1 \sim \mathcal{N}(0, I)$ and integrates backward to $t=0$ using a numerical solver such as Euler discretization:

\begin{equation}
x_{t+\Delta t} = x_t + \Delta t \, v_\theta(x_t, t, o).
\end{equation}

In practice, $N=8\sim10$ solver steps are used per control cycle to ensure action precision.
Each step requires a forward pass through the language backbone and policy head,
making the total inference cost scale linearly with $N$.
Under high frequency closed loop control, this iterative structure becomes a major latency bottleneck.

\subsection{Temporal Token Reuse}
As illustrated in Fig.~\ref{fig:token_reuse}, we adopt a simple odd even chunking strategy to identify tokens requiring updates. On odd chunks, we perform a full forward pass and cache key projections as reference representations. On even chunks, we compute cosine similarity between current and previously cached keys:
\begin{equation}
\text{sim}(i) = \cos(K_t^{(i)}, K_{\text{ref}}^{(i)}),
\end{equation} and select the top-$\rho N$ tokens with lowest similarity for update, where $\rho$ controls the update ratio.

As shown in Fig.~\ref{fig:overview}(a,b), token similarity patterns exhibit strong consistency across ViT layers. In particular, tokens that exhibit large changes at early layers tend to remain salient in deeper layers.
Thus, index selection is performed only in the first layer and reused across subsequent layers, significantly reducing overhead.

\begin{equation}
Q_t^{(\mathcal{U}_t)} = W_Q T_t^{(\mathcal{U}_t)}, \quad
K_t^{(\mathcal{U}_t)} = W_K T_t^{(\mathcal{U}_t)}, \quad
V_t^{(\mathcal{U}_t)} = W_V T_t^{(\mathcal{U}_t)}.
\end{equation} Updated tokens are recomputed, while others reuse cached key value pairs.
The full representation is assembled via sparse scatter:

\begin{equation}
K_t^{(\text{full})} = \text{scatter}(K_t^{(\mathcal{U}_t)}, K_{\text{ref}}^{(\mathcal{R}_t)}).
\end{equation} Attention is computed only for updated tokens, while reused tokens directly inherit cached outputs.
After residual addition, all tokens are processed by the MLP layer, ensuring full forward propagation and stable training. As illustrated in Fig.~\ref{fig:token_reuse}, the proposed mechanism selectively updates a subset of tokens while reusing the remaining ones across consecutive frames. Specifically, the first Transformer layer performs token selection based on cosine similarity, where tokens with large feature variations are identified as dynamic regions (top-$\rho N$), while the rest are treated as temporally redundant and skipped. The selected indices are then reused across subsequent layers, where only the corresponding tokens are recomputed, and all others directly inherit cached representations. This design is motivated by the empirical observation, discussed in Sec.1, that token similarity patterns remain highly consistent across layers.
\subsection{Efficient Flow Matching Policy with Step Compression}
\begin{table*}[t]
\centering
\caption{Comparison of VLA models on the LIBERO benchmark.
We report success rate (SR) and efficiency metrics.  SR for each column corresponds to individual task categories, while Mean SR denotes the average performance across all tasks.}
\label{tab:main_results}

\resizebox{0.98\textwidth}{!}{
\begin{tabular}{lccccccccc}
\toprule
\textbf{Method}
& \textbf{LIBERO 10}
& \textbf{Goal}
& \textbf{Spatial}
& \textbf{Object}
& \textbf{Mean SR}
& \textbf{Steps}
& \textbf{Avg Time (ms)}
& \textbf{FPS}
& \textbf{TFLOPs} \\

\midrule
$\pi_0$ \cite{black2024pi_0}
& 83.2 & 90.0 & 98.0 & 97.8
& 92.3
& 10 & 276.4 & 3.6 & 4.48 \\

$\pi_{0.5}$ \cite{intelligence2025pi_}
& 88.6 & 92.0 & 98.6 & 98.2
& 94.4
& 10 & 286.9 & 3.5 & 4.48 \\

OpenVLA \cite{kim2024openvla}
& 53.7 & 79.2 & 84.7 & 88.4
& 76.5
& 8 & 629.6 & 1.6 & 8.82 \\

X-VLA \cite{zheng2025x}
& 88.9 & 91.2 & 98.2 & 98.0
& 94.1
& 10 & 256.8 & 3.9 & -- \\

Efficient VLA \cite{yang2025efficientvla}
& 81.2 & 91.6 & 95.8 & 97.8
& 91.6
& 10 & 144.1 & 6.9 & 1.48 \\

\midrule

\rowcolor{blue!5}
\textbf{$\pi_{0}$ + Our Method}
& 80.2 & 89.6 & 96.8 & 97.2
& 91.0
& \textbf{2} & 156.2 & 6.4 & 2.86 \\

\rowcolor{blue!5}
\textbf{$\pi_{0.5}$ + Ours Method}
& 88.2 & 91.2 & 97.8 & 98.0
& 93.8
& \textbf{2} & \textbf{121.2} & \textbf{8.2} & 1.23 \\
\bottomrule
\bottomrule
\end{tabular}
}

\end{table*}
We analyze the intrinsic structure of the learned flow dynamics.
As shown in Fig.~\ref{fig:overview}(c), the singular value spectrum of the velocity sequence exhibits a rapid decay, with most of the energy concentrated in the top two components. Motivated by this observation, we model the velocity dynamics as a low rank process,
where the trajectory evolves within a low dimensional subspace.

To further validate this observation, we visualize the action trajectories in the principal component space in Fig.~\ref{fig:flow_compression}.
We observe that the trajectories produced by the compressed 2-step solver closely match those of the original 10-step solver in the top-2 principal components,
indicating that the trajectory evolution is intrinsically low dimensional and can be captured by a small number of degrees of freedom. Use $\{v_t\}_{t=0}^{T-1}$ denote the velocity sequence.
Under the low rank assumption, each velocity can be expressed as a linear combination of a small number of basis directions:\begin{equation}
v_t = \sum_{k=1}^{r} \alpha_{t,k} u_k + \epsilon_t.
\end{equation} In practice, we choose $v_0$ and $v_7$ as anchor velocities, as they respectively represent early stage global direction and mid trajectory refinement, 
providing a sufficiently good basis to capture the dominant dynamics under the low rank assumption. Namely the initial velocity $v_0$ and a representative mid trajectory velocity $v_7$, leading:
\begin{equation}
v_t \approx a_t v_0 + b_t v_7 + \epsilon_t.
\end{equation} Under Euler integration, the final state is given by:
\begin{equation}
x_T = x_0 + \sum_{t=0}^{T-1} \Delta t \cdot v_t.
\end{equation} Substituting the low rank approximation yields:
\begin{equation}
x_T \approx x_0 + \alpha v_0 + \beta v_7 + b,
\label{eq:two_anchor_velocity}
\end{equation} where $\alpha, \beta$ aggregate temporal contributions and $b$ absorbs residual errors. As illustrated in Fig.~\ref{fig:flow}, this formulation allows us to reconstruct the full trajectory using only two representative steps, effectively reducing the multi step solver to a compact 2-step formulation. The coefficients $(\alpha, \beta, b)$ depend on the observation and are not directly available in practice. To estimate these parameters, we introduce a lightweight adaptor $g_\phi$ that directly predicts:
\begin{equation}
(\alpha, \beta, b) = g_\phi(v_0).
\end{equation} Training the adaptor by minimizing the discrepancy between the reconstructed trajectory using the compressed 2-step formulation and the original full step trajectory in practice overall:
\begin{equation}
\label{eq:calib_loss}
\min_{\phi}\;
\mathbb{E}_{o \sim \mathcal{D}_{\text{cal}}}
\Big[
\| x_T^{(2)} - x_T^{(10)} \|_2^2
\Big].
\end{equation}
\section{Experiment}
\subsection{Experimental Setup}
\paragraph{Simulation Benchmark.}
We evaluate our framework on the LIBERO and RobotWin benchmark\cite{liu2023libero, chen2025robotwin}, these two widely used simulation suites for tabletop manipulation tasks. The model takes $224 \times 224$ RGB observations and natural language instructions as input and directly outputs joint space actions for 7-DoF manipulation. Success rate is adopted as the primary evaluation metric.
\paragraph{Real Robot Platform.}
To validate the proposed acceleration framework in real world settings, we deploy our method on the \textbf{Marvin Pro} robotic platform, a bimanual robot with two 7-DoF arms. 
Visual observations are provided by three RGB-D cameras: a head mounted Intel RealSense D435 and two wrist cameras SHW 5G. Demonstration data are collected via teleoperation using a Meta  Quest 3. The policy outputs actions in Cartesian space control.
\subsection{Simulation Results}

\begin{table*}[t]
\centering
\caption{
Inference time breakdown and computational cost on LIBERO and RoboTwin.
Numbers in parentheses indicate relative values normalized to $\pi_{0.5}$ and $\pi_{0}$ (100.0\%).
}
\label{tab:efficiency_breakdown_full}

\small

\renewcommand{\arraystretch}{1.2}

\centering{\textbf{LIBERO Benchmark}} \vspace{4pt}

\resizebox{\linewidth}{!}{
\begin{tabular}{lccccccc}
\toprule
Method & SR (\%) $\uparrow$ & ViT (ms) $\downarrow$ & LLM (ms) $\downarrow$ & Action Expert (ms) $\downarrow$ & Total (ms) $\downarrow$ & FPS $\uparrow$ & TFLOPs $\downarrow$ \\
\midrule
$\pi_{0.5}$ & 94.4 & 40.1 {\color{percentcolor}(100\%)} & 42.5 {\color{percentcolor}(100\%)} & 212.6 {\color{percentcolor}(100\%)} & 293.2 {\color{percentcolor}(100\%)} & 3.4 {\color{percentcolor}(100\%)} & 4.48 {\color{percentcolor}(100\%)} \\
ToMe\cite{bolya2022token} & 75.3 & 31.8 {\color{percentcolor}(79\%)} & 32.8 {\color{percentcolor}(77\%)} & 202.3 {\color{percentcolor}(95\%)} & 266.9 {\color{percentcolor}(91\%)} & 3.7 {\color{percentcolor}(109\%)} & 3.42 {\color{percentcolor}(76\%)} \\
ToFu\cite{maini2024tofu} & 71.6 & 31.2 {\color{percentcolor}(78\%)} & 30.1 {\color{percentcolor}(71\%)} & 204.2 {\color{percentcolor}(96\%)} & 265.5 {\color{percentcolor}(91\%)} & 3.8 {\color{percentcolor}(112\%)} & 3.25 {\color{percentcolor}(73\%)} \\
V2Drop\cite{chen2025variationawarevisiontokendropping} & 68.2 & 32.4 {\color{percentcolor}(81\%)} & 29.1 {\color{percentcolor}(68\%)} & 202.8 {\color{percentcolor}(95\%)} & 264.3 {\color{percentcolor}(90\%)} & 3.8 {\color{percentcolor}(112\%)} & 3.18 {\color{percentcolor}(71\%)} \\
SnapKV\cite{li2024snapkv} & 89.8 & 27.2 {\color{percentcolor}(68\%)} & 26.1 {\color{percentcolor}(61\%)} & 196.3 {\color{percentcolor}(92\%)} & 249.6 {\color{percentcolor}(85\%)} & 4.0 {\color{percentcolor}(118\%)} & 3.62 {\color{percentcolor}(81\%)} \\
SparseVLM\cite{zhang2024sparsevlm} & 92.2 & 30.2 {\color{percentcolor}(75\%)} & 31.1 {\color{percentcolor}(73\%)} & 199.6 {\color{percentcolor}(94\%)} & 260.9 {\color{percentcolor}(89\%)} & 3.8 {\color{percentcolor}(112\%)} & 3.85 {\color{percentcolor}(86\%)} \\
DART\cite{yin2025dart}
& 91.8
& 24.6 {\color{percentcolor}(61\%)}
& 22.8 {\color{percentcolor}(54\%)}
& 201.7 {\color{percentcolor}(95\%)}
& 249.1 {\color{percentcolor}(85\%)}
& 4.0 {\color{percentcolor}(118\%)}
& 3.41 {\color{percentcolor}(76\%)} \\
VLA-cache \cite{xu2025vla} & 90.2 & 28.2 {\color{percentcolor}(70\%)} & 20.2 {\color{percentcolor}(48\%)} & 198.2 {\color{percentcolor}(93\%)} & 246.6 {\color{percentcolor}(84\%)} & 4.1 {\color{percentcolor}(121\%)} & 3.57 {\color{percentcolor}(80\%)} \\
\midrule
\textbf{Ours} & \textbf{93.8} & \textbf{38.6 {\color{percentcolor}(96\%)}} & \textbf{42.3 {\color{percentcolor}(99\%)}} & \textbf{40.9 {\color{percentcolor}(19\%)}} & \textbf{121.8 {\color{percentcolor}(42\%)}} & \textbf{8.2 {\color{percentcolor}(241\%)}} & \textbf{1.23 {\color{percentcolor}(64\%)}} \\
\bottomrule
\end{tabular}
}

\vspace{18pt} 

\centering{\textbf{RoboTwin 2.0 Benchmark}} \vspace{4pt}

\resizebox{\linewidth}{!}{ 
\begin{tabular}{lccccccc}
\toprule
Method & TOP10 SR (\%) $\uparrow$ & ViT (ms) $\downarrow$ & LLM (ms) $\downarrow$ & Action Expert (ms) $\downarrow$ & Total (ms) $\downarrow$ & FPS $\uparrow$ & TFLOPs $\downarrow$ \\
\midrule
$\pi_{0}$ & 82.2 & 41.25 {\color{percentcolor}(100\%)} & 26.05 {\color{percentcolor}(100\%)} & 226.85 {\color{percentcolor}(100\%)} & 298.46 {\color{percentcolor}(100\%)} & 3.35 {\color{percentcolor}(100\%)} & 4.38 {\color{percentcolor}(100\%)} \\
ToMe\cite{bolya2022token} & 66.1 & 37.11 {\color{percentcolor}(90\%)} & 26.18 {\color{percentcolor}(101\%)} & 200.60 {\color{percentcolor}(88\%)} & 268.97 {\color{percentcolor}(90\%)} & 3.72 {\color{percentcolor}(111\%)} & 3.35 {\color{percentcolor}(76\%)} \\
ToFu\cite{wang2022multimodal} & 60.4 & 36.50 {\color{percentcolor}(88\%)} & 25.96 {\color{percentcolor}(100\%)} & 199.34 {\color{percentcolor}(88\%)} & 265.00 {\color{percentcolor}(89\%)} & 3.77 {\color{percentcolor}(113\%)} & 3.38 {\color{percentcolor}(77\%)} \\
V2Drop\cite{chen2025variationawarevisiontokendropping} & 59.8 & 36.03 {\color{percentcolor}(87\%)} & 26.15 {\color{percentcolor}(100\%)} & 198.25 {\color{percentcolor}(87\%)} & 264.70 {\color{percentcolor}(89\%)} & 3.78 {\color{percentcolor}(113\%)} & 3.60 {\color{percentcolor}(82\%)} \\
SnapKV\cite{li2024snapkv} & 78.3 & 34.31 {\color{percentcolor}(83\%)} & 25.68 {\color{percentcolor}(99\%)} & 188.30 {\color{percentcolor}(83\%)} & 252.87 {\color{percentcolor}(85\%)} & 3.95 {\color{percentcolor}(118\%)} & 3.90 {\color{percentcolor}(89\%)} \\
SparseVLM\cite{zhang2024sparsevlm} & 76.4 & 36.58 {\color{percentcolor}(89\%)} & 25.89 {\color{percentcolor}(99\%)} & 200.29 {\color{percentcolor}(88\%)} & 265.98 {\color{percentcolor}(89\%)} & 3.76 {\color{percentcolor}(112\%)} & 3.85 {\color{percentcolor}(88\%)} \\
\midrule
\textbf{Ours} & \textbf{81.5} 
& \textbf{39.6 {\color{percentcolor}(96\%)}} 
& \textbf{25.8 {\color{percentcolor}(99\%)}} 
& \textbf{43.1 {\color{percentcolor}(19\%)}} 
& \textbf{125.4 {\color{percentcolor}(42\%)}} 
& \textbf{8.0 {\color{percentcolor}(239\%)}} 
& \textbf{2.80 {\color{percentcolor}(64\%)}} \\
\bottomrule
\end{tabular}
}
\end{table*}
We evaluate our method on the LIBERO benchmark and compare it with representative VLA baselines.
As shown in Table~\ref{tab:main_results}, our approach consistently maintains competitive task performance while significantly reducing the number of sampling steps overall.

In particular, applying our method to $\pi_{0.5}$ achieves a mean success rate of 93.8\%, broadly comparable to the original $\pi_{0.5}$ and X-VLA, despite significantly reducing the sampling process from 10 steps to 2. This strongly suggests that a substantial portion of iterative updates in flow based policies may be largely redundant. Meanwhile, our method substantially improves efficiency, reducing inference latency from 286.9 ms to 121.2 ms while correspondingly increasing control frequency from 3.5 FPS to 8.2 FPS. Similar trends are observed for $\pi_0$, demonstrating that the proposed step compression strategy generalizes across different policy backbones.

Compared to prior methods that primarily optimize perception or language modules, our approach directly and effectively reduces the dominant cost in action generation, thereby ultimately leading to more substantial overall end to end speedup in practice.

Moreover, the efficiency gains are achieved without introducing additional model complexity or training overhead. Overall, these results clearly demonstrate a favorable trade off between efficiency and performance, making the approach suitable for real time robotic deployment without sacrificing task performance.

\subsection{Inference Efficiency Analysis}

We analyze the source of acceleration by decomposing the inference pipeline into perception (ViT), language reasoning (LLM), and action generation (Action Expert), as summarized in Table~\ref{tab:efficiency_breakdown_full}.

The results clearly show that, in the evaluated VLA models, the primary computational bottleneck lies in the Action Expert, which largely dominates the overall latency, while the ViT and LLM contribute a relatively much smaller portion overall.

Our method directly targets this bottleneck. While leaving the ViT and LLM latency largely unchanged, it significantly reduces the cost of the Action Expert, leading to a substantial reduction in end to end latency and a corresponding increase in control frequency.

This suggests that a major source of inefficiency originates from the multi step policy sampling process. In contrast, many existing efficiency approaches primarily focus on optimizing perception or language modules, which may lead to limited end to end gains when action generation dominates the computational cost.

Moreover, this finding highlights an important system level insight: optimizing upstream modules such as perception or language reasoning alone may yield limited overall acceleration when downstream policy execution remains the dominant bottleneck. In contrast, directly reducing the computational cost of action generation leads to more substantial end to end efficiency gains, in practice overall, emphasizing the importance of holistic optimization across the entire VLA pipeline. Overall, these results clearly highlight that policy level compression provides an effective strategy for improving system level efficiency in flow based VLA models, achieving both high efficiency and competitive performance in practice.

\subsection{System Level Ablation}
\begin{table*}
\centering
\caption{
\textbf{System level ablation of our method.}
We report success rate and efficiency on Libero. 
``TokenReuse'' enables temporal token reuse in the vision encoder, and ``2-Step'' compresses flow matching from 10 steps to 2 steps.
}
\label{tab:ablation_system}

\setlength{\tabcolsep}{5pt} 
\renewcommand{\arraystretch}{1.15}

\begin{tabular}{lccccccc}
\toprule
Method & TokenReuse & Efficient Policy 
& Mean SR (\%) 
& ViT (ms)$\downarrow$ & LLM (ms)$\downarrow$ & Action (ms)$\downarrow$
& Total (ms)$\downarrow$ / FPS$\uparrow$ \\
\midrule

Baseline ($\pi_{0.5}$, 10-step) & \xmark & \xmark 
& 94.4 & 40.1 & 42.5 & 212.6 & 293.2 / 3.4 \\

\midrule

Token Reuse & \cmark & \xmark
& 93.9 & 28.5 & 42.6 & 211.8 & 282.9 / 3.5 \\

Efficient Policy & \xmark & \cmark
& 94.1 & 40.0 & 42.4 & 41.5 & 123.9 / 8.1 \\

\midrule

\textbf{Ours} & \cmark & \cmark
& \textbf{93.8} & \textbf{28.4} & \textbf{42.3} & \textbf{40.9} & \textbf{111.6 / 9.0} \\

\bottomrule
\end{tabular}

\end{table*}

\begin{figure*}
\centering
\includegraphics[width=\textwidth]{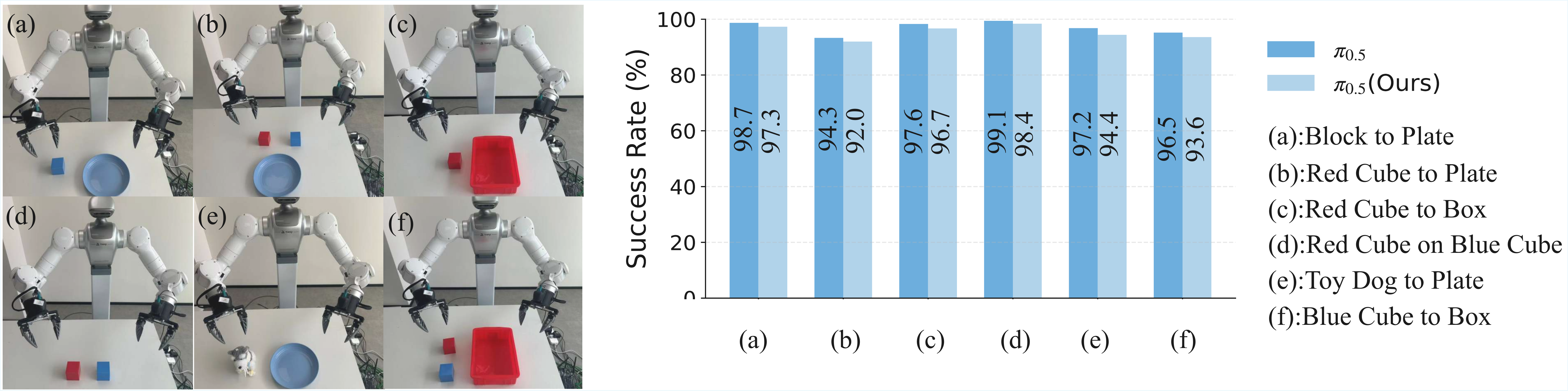}
\caption{
\textbf{Real world manipulation performance on the Marvin Pro robot.}
\textbf{Left:} Representative task snapshots from six manipulation tasks:
(a) put the blue block on the plate,
(b) place the red block onto the blue plate,
(c) put the red block into the box,
(d) stack the red block on top of the blue block,
(e) pick up the dog toy and place it on the plate,
(f) place the blue block into the red box.
\textbf{Right:} Task level success rate comparison between $\pi_{0.5}$ with the standard 10-step policy and our method.
}
\label{fig:real_world_results}
\end{figure*}

\begin{table*}[t]
\centering
\caption{
\textbf{Real robot evaluation.}
We report task success rate (SR) and success rate within 30 seconds (SR@30s) across six manipulation tasks.
}
\label{tab:real_robot}

\setlength{\tabcolsep}{5pt}
\renewcommand{\arraystretch}{1.15}

\begin{tabular*}{\textwidth}{@{\extracolsep{\fill}}>{\centering\arraybackslash}p{0.45\textwidth}cccc@{}}
\toprule
Task 
& \multicolumn{2}{c}{$\pi_{0.5}$} 
& \multicolumn{2}{c}{$\pi_{0.5}$ (Ours)} \\
\cmidrule(lr){2-3} \cmidrule(lr){4-5}
& SR (\%) & SR@30s (\%) 
& SR (\%) & SR@30s (\%) \\
\midrule

(a) Put the block on the plate 
& 98.7 {\color{percentcolor}(100\%)} 
& 78.7 {\color{percentcolor}(79.7\%)} 
& 97.3 {\color{percentcolor}(98.6\%)} 
& 89.3 {\color{percentcolor}(90.5\%)} \\

(b) Place the red cube onto the plate 
& 97.3 {\color{percentcolor}(100\%)} 
& 79.3 {\color{percentcolor}(81.5\%)} 
& 92.0 {\color{percentcolor}(94.6\%)} 
& 82.0 {\color{percentcolor}(84.3\%)} \\

(c) Place the red cube in the box 
& 98.3 {\color{percentcolor}(100\%)} 
& 81.3 {\color{percentcolor}(82.7\%)} 
& 96.7 {\color{percentcolor}(98.4\%)} 
& 91.7 {\color{percentcolor}(93.3\%)} \\

(d) Place the dog toy onto the plate 
& 99.4 {\color{percentcolor}(100\%)} 
& 68.4 {\color{percentcolor}(68.8\%)} 
& 98.4 {\color{percentcolor}(99.0\%)} 
& 75.4 {\color{percentcolor}(75.9\%)} \\

(e) Place the red cube onto the blue cube 
& 96.8 {\color{percentcolor}(100\%)} 
& 66.8 {\color{percentcolor}(69.0\%)} 
& 94.4 {\color{percentcolor}(97.5\%)} 
& 65.4 {\color{percentcolor}(67.6\%)} \\

(f) Place the blue cube in the red box 
& 95.2 {\color{percentcolor}(100\%)} 
& 88.2 {\color{percentcolor}(92.6\%)} 
& 93.6 {\color{percentcolor}(98.3\%)} 
& 90.6 {\color{percentcolor}(95.2\%)} \\

\midrule
\textbf{Mean}
& \textbf{97.2} {\color{percentcolor}(100\%)} 
& 77.1 {\color{percentcolor}(79.3\%)} 
& 95.4 {\color{percentcolor}(98.1\%)} 
& \textbf{82.3} {\color{percentcolor}(84.7\%)} \\
\bottomrule
\end{tabular*}
\end{table*}
We analyze the individual contribution of each component in Table~\ref{tab:ablation_system}, including temporal token reuse in the vision encoder and the proposed 2-step efficient policy. Enabling Token Reuse alone reduces the ViT latency from 40.1\,ms to 28.5\,ms with almost no performance degradation (93.9\% vs.\ 94.4\%), showing that substantial temporal redundancy exists in visual tokens across consecutive frames. Compared to Token Reuse, the Efficient Policy contributes the majority of the latency reduction, further confirming that action generation is the dominant source of computational cost.

Applying the Efficient Policy alone dramatically reduces the Action Expert latency from 212.6\,ms to 41.5\,ms, significantly lowering the total inference time from 293.2\,ms to 123.9\,ms while maintaining comparable task performance (94.1\%). This suggests that compressing the multi step flow matching process largely maintains action performance while reducing redundant policy updates.

Combining both components yields the full model, achieving the lowest latency (111.6\,ms, 9.0 FPS) with minimal performance drop (93.8\%). These results validate that two components target different parts of the pipeline and provide complementary gains, enabling system level acceleration without sacrificing task success rate.

\subsection{Real World Evaluation}
\begin{figure*}[t]
\centering
\includegraphics[width=\textwidth]{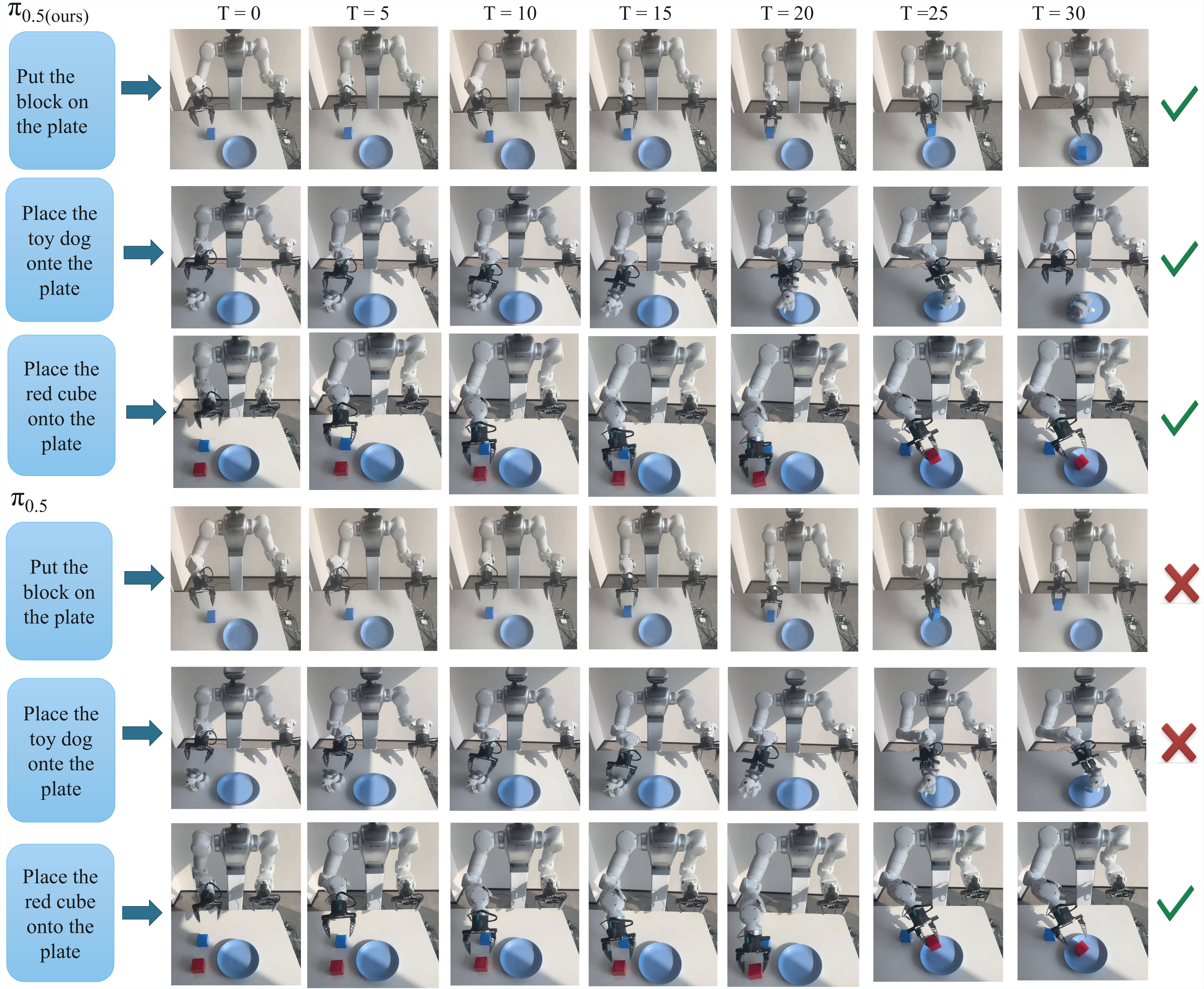}
\caption{
\textbf{Task rollout comparison under a fixed time budget.}
Each row shows a rollout trajectory from $T=0$ to $T=30$ for representative manipulation tasks. 
The top three rows correspond to $\pi_{0.5}$ with our 2-step compressed policy, while the bottom three rows show the original $\pi_{0.5}$ policy.
Each column visualizes the robot state at different timestamps during execution. The results show that our compressed policy consistently completes tasks within the time budget while preserving stable manipulation behavior. The results show that our compressed policy consistently completes tasks within the time budget while preserving stable manipulation behavior.}
\label{fig:policy_demo}
\end{figure*}

We further validate our method on a Marvin Pro Robot, as shown in Fig.~\ref{fig:real_world_results}. 
Across six representative tasks, our 2-step policy achieves a comparable overall success rate to the original $\pi_{0.5}$ (95.4\% vs.\ 97.2\%), while consistently improving the time constrained success rate from 77.1\% to 82.3\% (Table~\ref{tab:real_robot}).

This improvement clearly indicates that reducing policy latency increases the effective control frequency, allowing the robot to react more promptly to state deviations and complete long horizon tasks within the given time budget. Qualitatively (Fig.~\ref{fig:real_world_results}), the compressed policy consistently maintains stable and precise manipulation behavior across tasks requiring multi stage execution, without introducing additional failure modes.

Per task results suggest that the gains mainly arise from faster task completion rather than sacrificing accuracy, implying that step compression preserves the underlying action trajectory while removing redundant intermediate updates. This indicates that the essential geometric structure of the trajectory is maintained while unnecessary iterative refinements are effectively eliminated, leading to more efficient policy execution. Consistent observations from rollout trajectories show in Fig.~\ref{fig:policy_demo} confirm 2-step policy produces smooth and goal directed behaviors under time constrained execution. The resulting trajectories remain stable and coherent across different tasks, without introducing oscillations or degraded control precision, demonstrating that the compressed policy retains reliable dynamics despite significantly reduced inference steps.
\vspace{-5pt}
\section{Conclusion, Limitations and Future Work}

We presented a system level acceleration framework for Vision-Language-Action models that reduces temporal redundancy in both perception and policy inference. By combining temporally aware visual token reuse with a compressed 2-step flow matching policy, the proposed method significantly improves inference efficiency while maintaining competitive manipulation performance. Experiments on the LIBERO benchmark and real robot tasks demonstrate that our approach achieves a favorable balance between task success rate and real time control efficiency. Nevertheless, several limitations remain. The compressed policy is mainly validated on short horizon manipulation tasks, and its effectiveness for long horizon planning or highly dynamic environments requires further investigation. In future work, we will explore adaptive step scheduling strategies and extend the framework to more diverse tasks.

\clearpage
\bibliographystyle{plainnat}
\bibliography{references}

@article{ma2024survey,
  title={A survey on vision-language-action models for embodied ai},
  author={Ma, Yueen and Song, Zixing and Zhuang, Yuzheng and Hao, Jianye and King, Irwin},
  journal={arXiv preprint arXiv:2405.14093},
  year={2024}
}

@article{awadalla2023openflamingo,
  title={Openflamingo: An open-source framework for training large autoregressive vision-language models},
  author={Awadalla, Anas and Gao, Irena and Gardner, Josh and Hessel, Jack and Hanafy, Yusuf and Zhu, Wanrong and Marathe, Kalyani and Bitton, Yonatan and Gadre, Samir and Sagawa, Shiori and others},
  journal={arXiv preprint arXiv:2308.01390},
  year={2023}
}

@inproceedings{li2022blip,
  title={Blip: Bootstrapping language-image pre-training for unified vision-language understanding and generation},
  author={Li, Junnan and Li, Dongxu and Xiong, Caiming and Hoi, Steven},
  booktitle={International conference on machine learning},
  pages={12888--12900},
  year={2022},
  organization={PMLR}
}

@inproceedings{radford2021learning,
  title={Learning transferable visual models from natural language supervision},
  author={Radford, Alec and Kim, Jong Wook and Hallacy, Chris and Ramesh, Aditya and Goh, Gabriel and Agarwal, Sandhini and Sastry, Girish and Askell, Amanda and Mishkin, Pamela and Clark, Jack and others},
  booktitle={International conference on machine learning},
  pages={8748--8763},
  year={2021},
  organization={PmLR}
}

@article{yue2024deer,
  title={Deer-vla: Dynamic inference of multimodal large language models for efficient robot execution},
  author={Yue, Yang and Wang, Yulin and Kang, Bingyi and Han, Yizeng and Wang, Shenzhi and Song, Shiji and Feng, Jiashi and Huang, Gao},
  journal={Advances in Neural Information Processing Systems},
  volume={37},
  pages={56619--56643},
  year={2024}
}

@article{kim2024openvla,
  title={Openvla: An open-source vision-language-action model},
  author={Kim, Moo Jin and Pertsch, Karl and Karamcheti, Siddharth and Xiao, Ted and Balakrishna, Ashwin and Nair, Suraj and Rafailov, Rafael and Foster, Ethan and Lam, Grace and Sanketi, Pannag and others},
  journal={arXiv preprint arXiv:2406.09246},
  year={2024}
}

@article{black2024pi,
  title={pi0: A Vision-Language-Action Flow Model for General Robot Control},
  author={Black, K and others},
  journal={Robotics: Science and Systems XXI},
  year={2024}
}

@article{han2022survey,
  title={A survey on vision transformer},
  author={Han, Kai and Wang, Yunhe and Chen, Hanting and Chen, Xinghao and Guo, Jianyuan and Liu, Zhenhua and Tang, Yehui and Xiao, An and Xu, Chunjing and Xu, Yixing and others},
  journal={IEEE transactions on pattern analysis and machine intelligence},
  volume={45},
  number={1},
  pages={87--110},
  year={2022},
  publisher={IEEE}
}

@article{croitoru2023diffusion,
  title={Diffusion models in vision: A survey},
  author={Croitoru, Florinel-Alin and Hondru, Vlad and Ionescu, Radu Tudor and Shah, Mubarak},
  journal={IEEE transactions on pattern analysis and machine intelligence},
  volume={45},
  number={9},
  pages={10850--10869},
  year={2023},
  publisher={Ieee}
}

@article{lipman2022flow,
  title={Flow matching for generative modeling},
  author={Lipman, Yaron and Chen, Ricky TQ and Ben-Hamu, Heli and Nickel, Maximilian and Le, Matt},
  journal={arXiv preprint arXiv:2210.02747},
  year={2022}
}

@article{zhao2023learning,
  title={Learning fine-grained bimanual manipulation with low-cost hardware},
  author={Zhao, Tony Z and Kumar, Vikash and Levine, Sergey and Finn, Chelsea},
  journal={arXiv preprint arXiv:2304.13705},
  year={2023}
}

@article{fu2024mobile,
  title={Mobile aloha: Learning bimanual mobile manipulation with low-cost whole-body teleoperation},
  author={Fu, Zipeng and Zhao, Tony Z and Finn, Chelsea},
  journal={arXiv preprint arXiv:2401.02117},
  year={2024}
}

@inproceedings{zitkovich2023rt,
  title={Rt-2: Vision-language-action models transfer web knowledge to robotic control},
  author={Zitkovich, Brianna and Yu, Tianhe and Xu, Sichun and Xu, Peng and Xiao, Ted and Xia, Fei and Wu, Jialin and Wohlhart, Paul and Welker, Stefan and Wahid, Ayzaan and others},
  booktitle={Conference on Robot Learning},
  pages={2165--2183},
  year={2023},
  organization={PMLR}
}

@article{wen2025tinyvla,
  title={Tinyvla: Towards fast, data-efficient vision-language-action models for robotic manipulation},
  author={Wen, Junjie and Zhu, Yichen and Li, Jinming and Zhu, Minjie and Tang, Zhibin and Wu, Kun and Xu, Zhiyuan and Liu, Ning and Cheng, Ran and Shen, Chaomin and others},
  journal={IEEE Robotics and Automation Letters},
  year={2025},
  publisher={IEEE}
}

@article{black2024pi_0,
  title={pi0: A Vision-Language-Action Flow Model for General Robot Control},
  author={Black, Kevin and Brown, Noah and Driess, Danny and Esmail, Adnan and Equi, Michael and Finn, Chelsea and Fusai, Niccolo and Groom, Lachy and Hausman, Karol and Ichter, Brian and others},
  journal={arXiv preprint arXiv:2410.24164},
  year={2024}
}

@inproceedings{zhai2023sigmoid,
  title={Sigmoid loss for language image pre-training},
  author={Zhai, Xiaohua and Mustafa, Basil and Kolesnikov, Alexander and Beyer, Lucas},
  booktitle={Proceedings of the IEEE/CVF international conference on computer vision},
  pages={11975--11986},
  year={2023}
}

@article{sapkota2025vision,
  title={Vision-language-action models: Concepts, progress, applications and challenges},
  author={Sapkota, Ranjan and Cao, Yang and Roumeliotis, Konstantinos I and Karkee, Manoj},
  journal={arXiv preprint arXiv:2505.04769},
  year={2025}
}

@article{howard2017mobilenets,
  title={Mobilenets: Efficient convolutional neural networks for mobile vision applications},
  author={Howard, Andrew G and Zhu, Menglong and Chen, Bo and Kalenichenko, Dmitry and Wang, Weijun and Weyand, Tobias and Andreetto, Marco and Adam, Hartwig},
  journal={arXiv preprint arXiv:1704.04861},
  year={2017}
}

@inproceedings{zhang2018shufflenet,
  title={Shufflenet: An extremely efficient convolutional neural network for mobile devices},
  author={Zhang, Xiangyu and Zhou, Xinyu and Lin, Mengxiao and Sun, Jian},
  booktitle={Proceedings of the IEEE conference on computer vision and pattern recognition},
  pages={6848--6856},
  year={2018}
}

@article{dosovitskiy2020image,
  title={An image is worth 16x16 words: Transformers for image recognition at scale},
  author={Dosovitskiy, Alexey},
  journal={arXiv preprint arXiv:2010.11929},
  year={2020}
}

@inproceedings{liu2021swin,
  title={Swin transformer: Hierarchical vision transformer using shifted windows},
  author={Liu, Ze and Lin, Yutong and Cao, Yue and Hu, Han and Wei, Yixuan and Zhang, Zheng and Lin, Stephen and Guo, Baining},
  booktitle={Proceedings of the IEEE/CVF international conference on computer vision},
  pages={10012--10022},
  year={2021}
}

@article{yu2019review,
  title={A review of recurrent neural networks: LSTM cells and network architectures},
  author={Yu, Yong and Si, Xiaosheng and Hu, Changhua and Zhang, Jianxun},
  journal={Neural computation},
  volume={31},
  number={7},
  pages={1235--1270},
  year={2019},
  publisher={MIT Press One Rogers Street, Cambridge, MA 02142-1209, USA journals-info}
}

@article{janner2022planning,
  title={Planning with diffusion for flexible behavior synthesis},
  author={Janner, Michael and Du, Yilun and Tenenbaum, Joshua B and Levine, Sergey},
  journal={arXiv preprint arXiv:2205.09991},
  year={2022}
}

@article{chi2025diffusion,
  title={Diffusion policy: Visuomotor policy learning via action diffusion},
  author={Chi, Cheng and Xu, Zhenjia and Feng, Siyuan and Cousineau, Eric and Du, Yilun and Burchfiel, Benjamin and Tedrake, Russ and Song, Shuran},
  journal={The International Journal of Robotics Research},
  volume={44},
  number={10-11},
  pages={1684--1704},
  year={2025},
  publisher={Sage Publications Sage UK: London, England}
}

@article{pearce2023imitating,
  title={Imitating human behaviour with diffusion models},
  author={Pearce, Tim and Rashid, Tabish and Kanervisto, Anssi and Bignell, Dave and Sun, Mingfei and Georgescu, Raluca and Macua, Sergio Valcarcel and Tan, Shan Zheng and Momennejad, Ida and Hofmann, Katja and others},
  journal={arXiv preprint arXiv:2301.10677},
  year={2023}
}

@article{lu2025dpm,
  title={Dpm-solver++: Fast solver for guided sampling of diffusion probabilistic models},
  author={Lu, Cheng and Zhou, Yuhao and Bao, Fan and Chen, Jianfei and Li, Chongxuan and Zhu, Jun},
  journal={Machine Intelligence Research},
  pages={1--22},
  year={2025},
  publisher={Springer}
}

@article{salimans2022progressive,
  title={Progressive distillation for fast sampling of diffusion models},
  author={Salimans, Tim and Ho, Jonathan},
  journal={arXiv preprint arXiv:2202.00512},
  year={2022}
}

@article{liu2023libero,
  title={Libero: Benchmarking knowledge transfer for lifelong robot learning},
  author={Liu, Bo and Zhu, Yifeng and Gao, Chongkai and Feng, Yihao and Liu, Qiang and Zhu, Yuke and Stone, Peter},
  journal={Advances in Neural Information Processing Systems},
  volume={36},
  pages={44776--44791},
  year={2023}
}

@article{chen2025robotwin,
  title={Robotwin 2.0: A scalable data generator and benchmark with strong domain randomization for robust bimanual robotic manipulation},
  author={Chen, Tianxing and Chen, Zanxin and Chen, Baijun and Cai, Zijian and Liu, Yibin and Li, Zixuan and Liang, Qiwei and Lin, Xianliang and Ge, Yiheng and Gu, Zhenyu and others},
  journal={arXiv preprint arXiv:2506.18088},
  year={2025}
}

@article{bolya2022token,
  title={Token merging: Your vit but faster},
  author={Bolya, Daniel and Fu, Cheng-Yang and Dai, Xiaoliang and Zhang, Peizhao and Feichtenhofer, Christoph and Hoffman, Judy},
  journal={arXiv preprint arXiv:2210.09461},
  year={2022}
}

@article{maini2024tofu,
  title={Tofu: A task of fictitious unlearning for llms},
  author={Maini, Pratyush and Feng, Zhili and Schwarzschild, Avi and Lipton, Zachary C and Kolter, J Zico},
  journal={arXiv preprint arXiv:2401.06121},
  year={2024}
}

@misc{chen2025variationawarevisiontokendropping,
      title={Variation-aware Vision Token Dropping for Faster Large Vision-Language Models}, 
      author={Junjie Chen and Xuyang Liu and Zichen Wen and Yiyu Wang and Siteng Huang and Honggang Chen},
      year={2025},
      eprint={2509.01552},
      archivePrefix={arXiv},
      primaryClass={cs.CV},
      url={https://arxiv.org/abs/2509.01552}, 
}

@article{li2024snapkv,
  title={Snapkv: Llm knows what you are looking for before generation},
  author={Li, Yuhong and Huang, Yingbing and Yang, Bowen and Venkitesh, Bharat and Locatelli, Acyr and Ye, Hanchen and Cai, Tianle and Lewis, Patrick and Chen, Deming},
  journal={Advances in Neural Information Processing Systems},
  volume={37},
  pages={22947--22970},
  year={2024}
}

@article{zhang2024sparsevlm,
  title={Sparsevlm: Visual token sparsification for efficient vision-language model inference},
  author={Zhang, Yuan and Fan, Chun-Kai and Ma, Junpeng and Zheng, Wenzhao and Huang, Tao and Cheng, Kuan and Gudovskiy, Denis and Okuno, Tomoyuki and Nakata, Yohei and Keutzer, Kurt and others},
  journal={arXiv preprint arXiv:2410.04417},
  year={2024}
}

@article{xu2025vla,
  title={Vla-cache: Towards efficient vision-language-action model via adaptive token caching in robotic manipulation},
  author={Xu, Siyu and Wang, Yunke and Xia, Chenghao and Zhu, Dihao and Huang, Tao and Xu, Chang},
  journal={arXiv e-prints},
  pages={arXiv--2502},
  year={2025}
}

@article{tang2025vlash,
  title={Vlash: Real-time vlas via future-state-aware asynchronous inference},
  author={Tang, Jiaming and Sun, Yufei and Zhao, Yilong and Yang, Shang and Lin, Yujun and Zhang, Zhuoyang and Hou, James and Lu, Yao and Liu, Zhijian and Han, Song},
  journal={arXiv preprint arXiv:2512.01031},
  year={2025}
}

@article{tan2025think,
  title={Think twice, act once: Token-aware compression and action reuse for efficient inference in vision-language-action models},
  author={Tan, Xudong and Yang, Yaoxin and Ye, Peng and Zheng, Jialin and Bai, Bizhe and Wang, Xinyi and Hao, Jia and Chen, Tao},
  journal={arXiv preprint arXiv:2505.21200},
  year={2025}
}

@article{pertsch2025fast,
  title={Fast: Efficient action tokenization for vision-language-action models},
  author={Pertsch, Karl and Stachowicz, Kyle and Ichter, Brian and Driess, Danny and Nair, Suraj and Vuong, Quan and Mees, Oier and Finn, Chelsea and Levine, Sergey},
  journal={arXiv preprint arXiv:2501.09747},
  year={2025}
}

@article{zheng2025x,
  title={X-vla: Soft-prompted transformer as scalable cross-embodiment vision-language-action model},
  author={Zheng, Jinliang and Li, Jianxiong and Wang, Zhihao and Liu, Dongxiu and Kang, Xirui and Feng, Yuchun and Zheng, Yinan and Zou, Jiayin and Chen, Yilun and Zeng, Jia and others},
  journal={arXiv preprint arXiv:2510.10274},
  year={2025}
}

@article{yang2025efficientvla,
  title={Efficientvla: Training-free acceleration and compression for vision-language-action models},
  author={Yang, Yantai and Wang, Yuhao and Wen, Zichen and Zhongwei, Luo and Zou, Chang and Zhang, Zhipeng and Wen, Chuan and Zhang, Linfeng},
  journal={arXiv preprint arXiv:2506.10100},
  year={2025}
}

@article{intelligence2025pi_,
  title={{Pi-0.5: A Vision-Language-Action Model with Open-World Generalization}},
  author={Intelligence, Physical and Black, Kevin and Brown, Noah and Darpinian, James and Dhabalia, Karan and Driess, Danny and Esmail, Adnan and Equi, Michael and Finn, Chelsea and Fusai, Niccolo and others},
  journal={arXiv preprint arXiv:2504.16054},
  year={2025}
}

@article{yin2025dart,
  title={DART: Differentiable Dynamic Adaptive Region Tokenizer for Vision Foundation Models},
  author={Yin, Shicheng and Yin, Kaixuan and Liu, Yang and Chen, Weixing and Lin, Liang},
  journal={arXiv preprint arXiv:2506.10390},
  year={2025}
}

@inproceedings{wang2022multimodal,
  title={Multimodal token fusion for vision transformers},
  author={Wang, Yikai and Chen, Xinghao and Cao, Lele and Huang, Wenbing and Sun, Fuchun and Wang, Yunhe},
  booktitle={Proceedings of the IEEE/CVF conference on computer vision and pattern recognition},
  pages={12186--12195},
  year={2022}
}

\clearpage

\appendices

\section{Implementation Details}

\paragraph{Hardware Setup.}
All simulation experiments are conducted on a workstation equipped with an NVIDIA A100 GPU, while real-world deployment runs on a workstation equipped with an NVIDIA RTX 4090 GPU. Table~\ref{tab:hardware_setup} summarizes the hardware configurations used for simulation and real-world inference.

\begin{table}[h]
\centering
\caption{Hardware configuration for simulation and real-world inference.}
\label{tab:hardware_setup}
\small
\begin{tabular}{lcc}
\toprule
\textbf{Component} & \textbf{Simulation} & \textbf{Real Robot} \\
\midrule
GPU & NVIDIA A100 & NVIDIA RTX 4090 \\
CPU & Intel Xeon Platinum 8368 & AMD Ryzen 9 9950X3D \\
Memory & 80 GB & 48 GB \\
Framework & PyTorch & PyTorch \\
Precision & FP16 & FP16 \\
\bottomrule
\end{tabular}
\end{table}

\section{Efficiency Analysis}

We further analyze the efficiency of our framework from two aspects: 
(1) reducing the number of flow-matching solver steps in the policy network, and 
(2) applying temporal token reuse in the visual encoder.

\paragraph{Effect of Flow Matching Solver Steps.}

We first analyze the inference speedup obtained by reducing the number of flow matching solver steps. 
Let $N$ denote the number of solver steps used during inference. 
Since each step requires a forward pass through the policy network, the inference latency approximately scales linearly with $N$.

Table~\ref{tab:step_scaling} reports the latency and relative speedup under different solver step configurations. 
The baseline corresponds to the original policy using $N=10$ steps.

\begin{table}[h]
\centering
\caption{Inference latency and speedup under different solver step numbers $N$.}
\label{tab:step_scaling}
\small
\begin{tabular}{ccc}
\toprule
\textbf{Steps ($N$)} & \textbf{Latency (ms)} & \textbf{Speedup} \\
\midrule
10 (baseline) & 286.9 & $1.0\times$ \\
8  & 230.1 & $1.25\times$ \\
6  & 178.5 & $1.61\times$ \\
4  & 145.7 & $1.97\times$ \\
2 (ours) & 121.2 & $2.37\times$ \\
\bottomrule
\end{tabular}
\end{table}

As expected, reducing the solver steps leads to substantial latency reduction. 
However, naive step reduction often causes noticeable degradation in task success rate due to inaccurate trajectory integration. 
Our proposed 2-step compressed solver preserves the dominant flow trajectory through the low-rank velocity approximation described in Sec.~3.3, enabling aggressive step reduction while maintaining stable manipulation performance.

\paragraph{Effect of Temporal Token Reuse Ratios.}
\begin{figure}
\centering
\includegraphics[width=\linewidth]{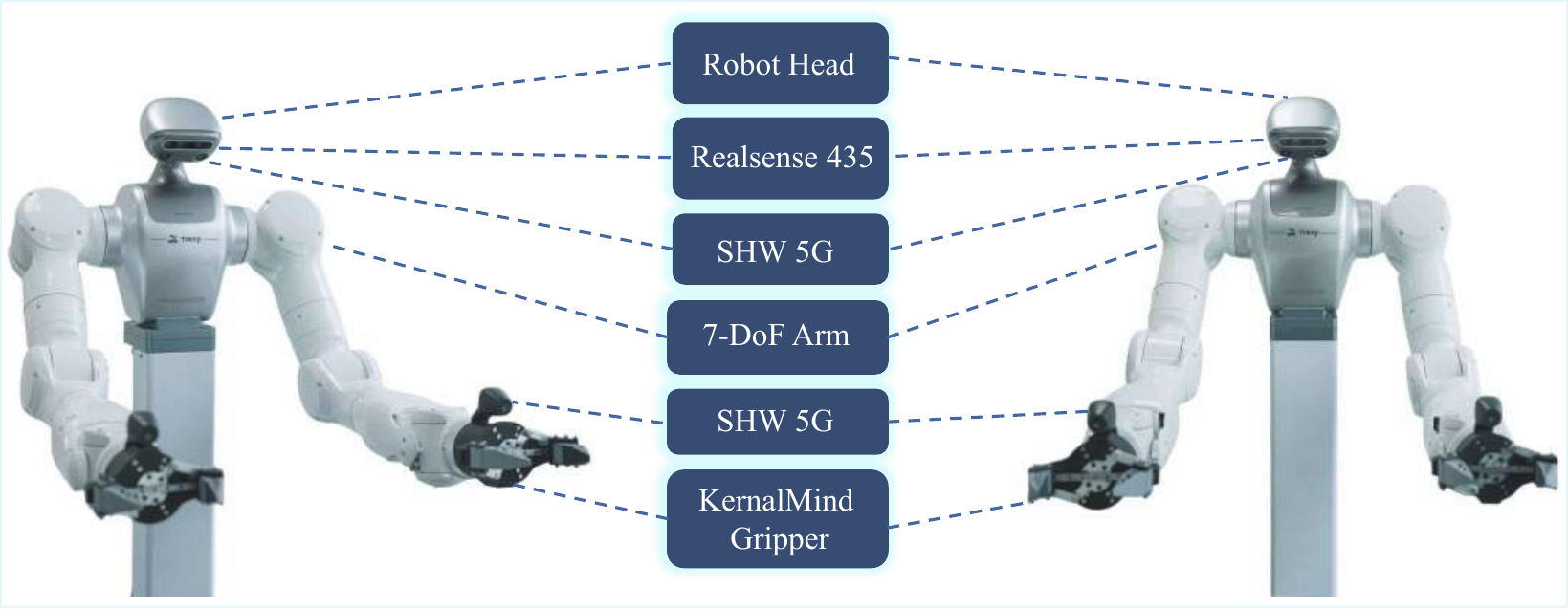}
\caption{
\textbf{Real robot platform.}
Dual-arm robot with 7-DoF manipulators, SHW 5G wrist modules, and KernalMind grippers.
A RealSense 435 camera is mounted on the head for visual input.
}
\label{fig:robot_system}
\end{figure}
We further analyze the effect of different token reuse ratios in the visual encoder. 
Let $r$ denote the proportion of tokens that are recomputed at each timestep, while the remaining tokens are reused from the previous frame.

Table~\ref{tab:token_ratio} reports the task success rate under different token update ratios. 
We observe that reducing the token update ratio slightly affects perception fidelity, which may lead to a gradual decrease in task success rate. 
However, moderate reuse ratios still maintain strong performance, indicating that most visual tokens remain temporally stable across adjacent frames.

In practice, we find that updating approximately $40\%$ of the tokens provides a good balance between visual representation accuracy and computational efficiency, and this configuration is adopted in our final system.

\begin{table}[h]
\centering
\caption{Effect of different token update ratios $r$ on task success rate.}
\label{tab:token_ratio}
\small
\begin{tabular}{cc}
\toprule
\textbf{Update Ratio ($r$)} & \textbf{Mean SR (\%)} \\
\midrule
1.0 (baseline) & 94.4 \\
0.2 & 94.1 \\
0.3 & 94.0 \\
0.4 & 93.8 \\
0.5 & 92.6 \\
0.6 & 91.8 \\
\bottomrule
\end{tabular}
\end{table}

\section{Full RoboTwin Benchmark Results.}

We further report detailed evaluation results on the RoboTwin benchmark, which contains a diverse set of manipulation tasks covering object relocation, stacking, and interaction scenarios.
As shown in Table~\ref{tab:sr_full}, our method achieves competitive success rates across a wide range of RoboTwin tasks, maintaining performance comparable to the baseline while outperforming other acceleration methods.

In addition, Table~\ref{tab:latency_full} demonstrates that our approach significantly reduces inference latency, achieving more than 2$\times$ speedup compared to the base model.
\begin{table*}[t]
\centering
\setlength{\tabcolsep}{5pt}

\begin{tabular*}{\textwidth}{@{\extracolsep{\fill}}
>{\centering\arraybackslash}p{0.30\textwidth}
ccccccc@{}}
\toprule
Task & Base & Ours & ToMe & ToFu & V2Drop & SnapKV & SparseVLM  \\
\midrule
adjust\_bottle & 85 & 72 & 70 & 64 & 62 & 81 & 81 \\
beat\_block\_hammer & 76 & 74 & 60 & 60 & 2 & 18 & 60 \\
blocks\_ranking\_rgb & 46 & 36 & 40 & 32 & 10 & 0 & 36 \\
blocks\_ranking\_size & 6 & 2 & 12 & 14 & 10 & 0 & 13 \\
click\_alarmclock & 68 & 48 & 52 & 48 & 46 & 64 & 64 \\
click\_bell & 40 & 34 & 38 & 38 & 18 & 30 & 38 \\
dump\_bin\_bigbin & 84 & 88 & 68 & 62 & 62 & 80 & 80 \\
grab\_roller & 92 & 94 & 76 & 70 & 68 & 88 & 88 \\
handover\_block & 50 & 52 & 16 & 6 & 30 & 0 & 11 \\
handover\_mic & 87 & 92 & 68 & 62 & 64 & 83 & 83 \\
hanging\_mug & 6 & 10 & 0 & 0 & 0 & 0 & 0 \\
lift\_pot & 52 & 68 & 46 & 44 & 10 & 38 & 45 \\
move\_can\_pot & 60 & 56 & 32 & 16 & 16 & 10 & 24 \\
move\_pillbottle\_pad & 32 & 30 & 20 & 20 & 12 & 14 & 20 \\
move\_playingcard\_away & 52 & 28 & 24 & 16 & 12 & 20 & 20 \\
move\_stapler\_pad & 0 & 0 & 0 & 0 & 0 & 0 & 0 \\
open\_laptop & 80 & 36 & 50 & 56 & 42 & 28 & 53 \\
open\_microwave & 72 & 66 & 78 & 78 & 26 & 0 & 78 \\
pick\_diverse\_bottles & 40 & 56 & 38 & 38 & 6 & 6 & 38 \\
pick\_dual\_bottles & 58 & 70 & 58 & 54 & 28 & 12 & 56 \\
place\_a2b\_left & 26 & 18 & 20 & 16 & 14 & 8 & 18 \\
place\_a2b\_right & 10 & 10 & 14 & 16 & 2 & 4 & 15 \\
place\_bread\_basket & 38 & 32 & 34 & 34 & 22 & 16 & 34 \\
place\_bread\_skillet & 26 & 40 & 24 & 22 & 12 & 10 & 23 \\
place\_burger\_fries & 68 & 80 & 55 & 50 & 46 & 64 & 65 \\
place\_can\_basket & 50 & 48 & 50 & 48 & 32 & 42 & 49 \\
place\_cans\_plasticbox & 30 & 16 & 10 & 16 & 20 & 0 & 13 \\
place\_container\_plate & 90 & 89 & 72 & 66 & 66 & 86 & 86 \\
place\_dual\_shoes & 28 & 22 & 20 & 14 & 6 & 0 & 17 \\
place\_empty\_cup & 60 & 56 & 46 & 42 & 44 & 57 & 57 \\
place\_fan & 10 & 0 & 0 & 0 & 0 & 8 & 0 \\
place\_mouse\_pad & 8 & 6 & 0 & 0 & 0 & 0 & 0 \\
place\_object\_basket & 24 & 48 & 20 & 8 & 2 & 0 & 14 \\
place\_object\_scale & 10 & 12 & 10 & 6 & 0 & 4 & 8 \\
place\_object\_stand & 68 & 58 & 58 & 36 & 32 & 24 & 47 \\
place\_phone\_stand & 30 & 32 & 26 & 28 & 8 & 4 & 27 \\
place\_shoe & 48 & 46 & 40 & 38 & 22 & 18 & 39 \\
press\_stapler & 44 & 58 & 20 & 16 & 20 & 28 & 18 \\
put\_bottles\_dustbin & 16 & 24 & 8 & 0 & 0 & 0 & 4 \\
put\_object\_cabinet & 30 & 22 & 18 & 16 & 4 & 0 & 17 \\
rotate\_qrcode & 56 & 56 & 50 & 36 & 22 & 4 & 43 \\
scan\_object & 22 & 22 & 20 & 8 & 6 & 10 & 14 \\
shake\_bottle\_horizontally & 96 & 100 & 80 & 72 & 72 & 92 & 92 \\
shake\_bottle & 92 & 96 & 74 & 68 & 68 & 88 & 88 \\
stack\_blocks\_three & 32 & 32 & 38 & 28 & 0 & 10 & 33 \\
stack\_blocks\_two & 62 & 62 & 48 & 46 & 42 & 28 & 47 \\
stack\_bowls\_three & 44 & 56 & 16 & 8 & 0 & 6 & 12 \\
stack\_bowls\_two & 90 & 90 & 68 & 70 & 50 & 36 & 39 \\
stamp\_seal & 16 & 36 & 18 & 20 & 2 & 16 & 19 \\
turn\_switch & 24 & 26 & 22 & 8 & 0 & 20 & 15 \\
\midrule
Average & 46.68 & 46.10 & 36.50 & 32.28 & 22.76 & 25.10 & 36.82 \\
\bottomrule
\end{tabular*}

\caption{Success rate comparison across tasks under different acceleration methods.}
\label{tab:sr_full}
\end{table*}
\begin{table*}[t]
\centering
\setlength{\tabcolsep}{5pt}

\begin{tabular*}{\textwidth}{@{\extracolsep{\fill}}
>{\centering\arraybackslash}p{0.30\textwidth}
ccccccc@{}}
\toprule
Task & Base & Ours & ToMe & ToFu & V2Drop & SnapKV & SparseVLM  \\
\midrule
adjust\_bottle & 295.4 & 124.1 & 266.2 & 262.3 & 261.9 & 250.3 & 263.2   \\
beat\_block\_hammer & 313.1 & 132.1 & 283.3 & 279.3 & 278.9 & 266.9 & 280.1 \\
blocks\_ranking\_rgb & 321.5 & 135.1 & 290.8 & 286.7 & 286.3 & 274.1 & 287.6 \\
blocks\_ranking\_size & 321.2 & 134.3 & 290.6 & 286.4 & 286.1 & 273.8 & 287.3 \\
click\_alarmclock & 285.6 & 120.0 & 257.4 & 253.5 & 253.2 & 241.9 & 254.4 \\
click\_bell & 295.1 & 124.0 & 267.0 & 263.2 & 262.9 & 251.6 & 264.1 \\
dump\_bin\_bigbin & 305.8 & 128.5 & 275.6 & 271.5 & 271.2 & 259.0 & 272.4 \\
grab\_roller & 298.2 & 125.3 & 268.7 & 264.7 & 264.4 & 252.6 & 265.6 \\
handover\_block & 307.5 & 129.4 & 278.2 & 274.2 & 273.9 & 262.2 & 275.1 \\
handover\_mic & 299.1 & 125.7 & 269.5 & 265.5 & 265.2 & 253.4 & 266.4 \\
hanging\_mug & 319.4 & 133.8 & 288.9 & 284.8 & 284.5 & 272.3 & 285.7 \\
lift\_pot & 312.2 & 131.0 & 282.4 & 278.5 & 278.1 & 266.2 & 279.3 \\
move\_can\_pot & 308.8 & 129.7 & 279.4 & 275.4 & 275.1 & 263.3 & 276.2 \\
move\_pillbottle\_pad & 307.4 & 128.7 & 278.1 & 274.1 & 273.8 & 262.1 & 275.0 \\
move\_playingcard\_away & 304.2 & 127.5 & 275.2 & 271.3 & 271.0 & 259.4 & 272.2 \\
move\_stapler\_pad & 307.1 & 128.6 & 277.8 & 273.9 & 273.6 & 261.8 & 274.7 \\
open\_laptop & 315.5 & 132.8 & 285.4 & 281.3 & 281.0 & 268.9 & 282.2 \\
open\_microwave & 317.7 & 133.1 & 287.4 & 283.3 & 283.0 & 270.8 & 284.2 \\
pick\_diverse\_bottles & 313.2 & 131.7 & 283.4 & 279.4 & 279.0 & 267.0 & 280.2 \\
pick\_dual\_bottles & 314.5 & 131.5 & 284.5 & 280.5 & 280.1 & 268.1 & 281.3 \\
place\_a2b\_left & 305.2 & 128.3 & 276.1 & 272.2 & 271.9 & 260.2 & 273.1 \\
place\_a2b\_right & 305.5 & 127.7 & 276.4 & 272.5 & 272.2 & 260.5 & 273.3 \\
place\_bread\_basket & 309.8 & 130.0 & 280.3 & 276.3 & 276.0 & 264.1 & 277.1 \\
place\_bread\_skillet & 311.2 & 131.1 & 281.5 & 277.5 & 277.2 & 265.3 & 278.4 \\
place\_burger\_fries & 312.4 & 131.3 & 281.5 & 277.4 & 277.0 & 264.6 & 278.3 \\
place\_can\_basket & 308.2 & 129.5 & 278.8 & 274.9 & 274.6 & 262.8 & 275.7 \\
place\_cans\_plasticbox & 310.5 & 130.3 & 280.9 & 276.9 & 276.6 & 264.7 & 277.8 \\
place\_container\_plate & 301.6 & 126.7 & 271.8 & 267.8 & 267.5 & 255.5 & 268.6 \\
place\_dual\_shoes & 316.2 & 133.4 & 286.0 & 282.0 & 281.7 & 269.5 & 282.8 \\
place\_empty\_cup & 292.4 & 122.8 & 263.5 & 259.6 & 259.3 & 247.7 & 260.5 \\
place\_fan & 315.8 & 132.2 & 285.7 & 281.6 & 281.3 & 269.2 & 282.5 \\
place\_mouse\_pad & 302.5 & 126.5 & 273.7 & 269.8 & 269.5 & 257.9 & 270.6 \\
place\_object\_basket & 309.7 & 130.6 & 280.2 & 276.2 & 275.9 & 264.0 & 277.1 \\
place\_object\_scale & 311.4 & 131.0 & 281.7 & 277.7 & 277.4 & 265.5 & 278.6 \\
place\_object\_stand & 311.9 & 131.4 & 282.2 & 278.1 & 277.8 & 265.9 & 279.0 \\
place\_phone\_stand & 306.5 & 128.3 & 277.3 & 273.3 & 273.1 & 261.3 & 274.2 \\
place\_shoe & 305.8 & 128.4 & 276.7 & 272.7 & 272.4 & 260.7 & 273.6 \\
press\_stapler & 301.1 & 126.8 & 272.4 & 268.5 & 268.2 & 256.7 & 269.4 \\
put\_bottles\_dustbin & 318.2 & 134.3 & 287.8 & 283.7 & 283.4 & 271.2 & 284.6 \\
put\_object\_cabinet & 320.5 & 135.2 & 289.9 & 285.8 & 285.5 & 273.2 & 286.7 \\
rotate\_qrcode & 299.4 & 125.2 & 270.9 & 267.0 & 266.7 & 255.3 & 267.9 \\
scan\_object & 300.2 & 126.4 & 271.6 & 267.7 & 267.4 & 256.0 & 268.6 \\
shake\_bottle\_horizontally & 297.4 & 125.0 & 268.0 & 264.0 & 263.7 & 251.9 & 264.9 \\
shake\_bottle & 296.8 & 124.6 & 267.5 & 263.5 & 263.2 & 251.4 & 264.4 \\
stack\_blocks\_three & 326.6 & 137.8 & 295.4 & 291.2 & 290.8 & 278.3 & 292.1 \\
stack\_blocks\_two & 321.4 & 135.1 & 290.7 & 286.6 & 286.3 & 273.9 & 287.5 \\
stack\_bowls\_three & 329.2 & 138.0 & 297.8 & 293.5 & 293.2 & 280.5 & 294.5 \\
stack\_bowls\_two & 325.1 & 136.2 & 294.1 & 289.9 & 289.5 & 277.1 & 290.8 \\
stamp\_seal & 302.8 & 127.7 & 274.0 & 270.0 & 269.7 & 258.2 & 270.9 \\
turn\_switch & 297.4 & 125.0 & 269.1 & 265.2 & 264.9 & 253.6 & 266.1 \\
\midrule
Avg & 309.30 & 129.67 & 278.37 & 274.28 & 273.95 & 262.86 & 275.32  \\
\bottomrule
\end{tabular*}
\caption{Average inference latency (ms) across RoboTwin tasks under different acceleration methods.}
\label{tab:latency_full}
\end{table*}

\end{document}